\documentclass{article} % For LaTeX2e
\usepackage{iclr2019_conference,times}

\usepackage{amsmath,amsfonts,bm}

\def\eqref#1{equation~\ref{#1}}
\def\1{\bm{1}}

\DeclareMathAlphabet{\mathsfit}{\encodingdefault}{\sfdefault}{m}{sl}
\SetMathAlphabet{\mathsfit}{bold}{\encodingdefault}{\sfdefault}{bx}{n}

\usepackage{hyperref}
\usepackage{url}
\usepackage{graphicx}

\title{Extractive summary as discrete latent variables}

\author{Aran Komatsuzaki\thanks{\url{https://github.com/AranKomat}} \\
School of Mathematics\\
Georgia Institute of Technology\\
Atlanta, GA 30332, USA \\
\texttt{akomatsuzaki3@gatech.edu} \\
}

\iclrfinalcopy % Uncomment for camera-ready version, but NOT for submission.
\begin{document}

\maketitle

\begin{abstract}
In this paper, we compare various methods to compress a text using a neural model. We find that extracting tokens as latent variables significantly outperforms the state-of-the-art discrete latent variable models such as VQ-VAE. Furthermore, we compare various extractive compression schemes. There are two best-performing methods that perform equally. One method is to simply choose the tokens with the highest tf-idf scores. Another is to train a bidirectional language model similar to ELMo and choose the tokens with the highest loss. If we consider any subsequence of a text to be a text in a broader sense, we conclude that language is a strong compression code of itself. Our finding justifies the high quality of generation achieved with hierarchical method, as their latent variables are nothing but natural language summary. We also conclude that there is a hierarchy in language such that an entire text can be predicted much more easily based on a sequence of a small number of keywords, which can be easily found by classical methods as tf-idf. We speculate that this extraction process may be useful for unsupervised hierarchical text generation.
\end{abstract}

\section{Introduction}
\label{intro}
For various sequence generation tasks, the performance of the state-of-the-art models is rapidly approaching to human-parity. Human-parity was achieved for an English to Chinese translation task \citep{parity}, and so were some other language pairs. In \citep{hier}, stories of about 800 word-length with high quality were generated, and \citep{wiki} achieved abstractive summarization of long documents with high quality. By regarding image generation as sequence generation with certain exploitable structure, \citep{spn} achieved generating images with an unprecedented quality and diversity. However, it is still difficult to perform unsupervised sequence generation of high quality that is unconditional or conditioned on a source sequence that contains small amount of information of the target sequence. Tasks such as translation and summarization are easier, since the source sequence contains sufficient information to construct the target sequence. Though the story generation in \citep{hier} is harder in this sense, it is still supervised; in other words, it requires the pairs of summary and original sequences, whose ample availability one cannot generally assume. 

To make the task unsupervised, one has to first generate a shorter sequence from a longer, original sequence by some compression method. From the spectacular result of \citep{hier} that significantly outperformed that of unconditional LM, we believe this approach, an unsupervised version of \citep{hier}, is worth considering. Notably, VQ-VAE is a state-of-the-art method that can, in particular, learn to generate in the both directions in the way to minimize the perplexity of short-to-long mapping \citep{music,vqvae,kaiser1,kaiser2}. While VQ-VAE, equipped with knowledge distillation, achieves non-autoregressive translation with BLEU close to the autoregressive state-of-the-art \citep{roy}, we show that, according to the metric of \citep{kaiser1}, the compressivity of VQ-VAE is significantly lower than that of some simple extraction methods we propose.

\section{Background / Related Works}
The fundamental motivation for our work is to improve the current state-of-the-art of sequence generation. In the following a few sections, we review the prominent approaches to text generation. 

\subsection{Non-MLE-based sequence generation}
There are various non-MLE based sequence generation, and the most popular approach is through GAN \citep{seqgan,leakgan,maskgan,rnn,wgan-gp,feature,texy}. \citep{eval} demonstrated that no GAN-based text generation model so far has outperformed LM. It was demonstrated that the BLEU and its variants are unreliable for unconditional text generation, and that the use of these metrics resulted in the proliferation of GAN models in this field. CoT is a sequence generation model based on cooperative training, whose training is similar to that of GAN but more stable \citep{cot}. Unlike GAN models, CoT achieved a better test perplexity than LM. However, we verified that LM outperforms CoT when we optimize their performance by allowing architectures other than LSTM such as Transformer, which dramatically boosts the performance of LM. There are also many VAE-based models \citep{control,cont}; however, we are not aware of any model that can perform better than LM does. 

There are also RL-based models without relying on GAN or VAE. One approach is to sample tokens using RL (or a simple heuristic) involving a pre-trained LM. However, it is difficult to outperform simple beam search ($k=5$) in this setting. For example, it is well-known that optimizing for the metrics such as BLEU does not necessarily result in improved quality \citep{google}. \citep{uncertainty} demonstrated that it is unlikely to improve the quality of generation by refining search algorithm to maximize cumulative probability, such as beam search, due to the mismatch between the prediction by LM and the real distribution.

There is also an approach through training a RL model from scratch or a RL model from a pre-trained LM. Both approaches suffer from poor scalability for the obvious reason that most RL algorithms are more prone to overfit and more sensitive to hyperparameters than LM. Note that there are some recently invented RL algorithms with less overfitting or sensitivity, such as \citep{alphazero}. We replaced the generator of SeqGAN with Alpha Zero, but we observed poor performance and severe overfitting. We encourage the future works on text generation to employ large datasets for meaningful evaluation.

\subsection{MLE-based sequence generation}
Given the insurmountable barrier in performance between MLE-based text generation and other methods, we believe the most promising approach is to stick with the MLE-based approach. Even within MLE training paradigm, there are many ways to proceed.  

\begin{figure*}[ht]
\vskip 0.2in
\begin{center}
\centerline{\includegraphics[width=\linewidth]{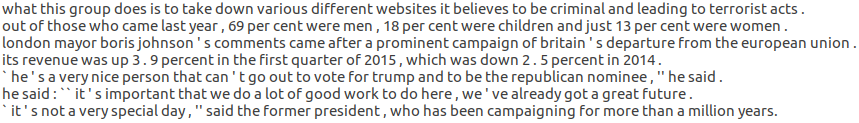}}
\caption{Samples generated by Transformer LM with temperature-tuned decoding}
\label{samples}
\end{center}
\vskip -0.2in
\end{figure*}
As far as we are aware, text generation with the state-of-the-art quality-diversity trade-off can be achieved with vanilla Transformer or Transformer-XL with beam search (for translation and chatbot) or temperature-tuned decoding, depending on the task of interest \citep{hier,falling}. The sampling method is crucial for better text generation. \citep{hier} used LM top-k sampling ($k=10$) and temperature of $\alpha=0.8$ (both for vanilla LM case and hierarchical LM case), which they found "substantially more effective than beam search, which tends to produce common phrases and repetitive text from the training set." \citep{falling} attempted vanilla LM text generation with temperature of $\alpha=0.7$. In either case, the quality-diversity trade-off of the generation is substantially better than any previous method. If the vocabulary of dataset is large, adaptive softmax and adaptive embedding can be used to dramatically reduce the number of parameters \citep{adap}. \citep{xl} showed that the use of recurrence achieved a better perplexity even on a dataset without long-range dependency such as 1BLM. 

Figure \ref{samples} shows some of the top 50\% of the samples generated by Transformer LM with temperature-tuned decoding. As the figure shows, the vanilla Transformer with the proper decoding generates with local dependency at a satisfactory level, whereas the global dependency is still weak. For example, the last sample is clearly unrealistic, given that a person has been campaigning for more than a million years. Since the local dependency is good even at the later part of a sentence, the weak global coherence is presumably not due to exposure bias. We believe that it is due to shortness of attention, which was identified as a problem for attention mechanism in \citep{frust}. This is in contrast with BERT \citep{mask}, which does not pose the shortness of attention. We believe this is because the bidirectional problems such as BERT and ELMo are substantially easier than unidirectional LM and therefore long-range attention can be established without special care. As an improvement on 1BLM was achieved with Transformer-XL, an improvement in the recurrent training in the sense of Transformer-XL may extend the span of attention, since it discourages for the gradient to flow in the short span. 

We should also note that a large dataset size is crucial for better quality of generation with LM. \citep{scaling} showed that each time doubling the dataset size and increasing the parameter size accordingly, one can reduce the perplexity by $4.5\%$ in 1BLM \citep{1b} without a sign of diminishing return as far as explored. For the case of text generation of general subjects, it is easy to collect billions of words from Internet.

To conclude, there are not many options we believe that are promising. In this paper, we do not investigate the recurrent training. Instead, we focus on the hierarchical approach taken by \citep{hier}. This approach reduces the problem of sequence generation by reducing it to Seq2Seq from a shorter sequence to a longer sequence. The shorter sequence can be generated from an even shorter sequence with Seq2Seq \citep{music} or from none with LM \citep{hier}. For the reason stated in Section \ref{intro}, we investigate the unsupervised compression of sequence.

\subsection{Compression of sequence}

\citep{kaiser1} proposed Improved Semantic Hashing (ISH), an autoencoder that compresses an input sequence into discrete latent variables and expands them for sequence generation, as well as DSAE, a metric to measure the rate of compression achieved by discrete latent variable model. At the moment of publication of \citep{kaiser1}, ISH was by far the best-performing discrete latent variable model. Later, \citep{roy} showed that VQ-VAE, equipped with knowledge distillation, achieves  significantly better BLEU in translation than ISH, which suggests VQ-VAE may achieve better DSAE than ISH. In our work, we show that our models significantly outperform ISH and VQ-VAE, both of which have roughly comparable DSAE, in term of DSAE. 

There is a work on achieving the state-of-the-art compression rate of a sentence by extracting the tokens \citep{compr}. Our work is different in the following ways. First, they aimed for a human-readable compression, whereas our aim is to achieve as low perplexity as possible for a given budget. We believe human-readability degrades the compression rate, since most human-readable text has to contain many tokens that are easy to predict given the rest of the sentence, such as prepositions, articles and other frequently appearing vocabulary to maintain correct grammar. Their aim aligns with that of the abstractive summarization, whereas ours is to achieve better hierarchical text generation and construct general-purpose discrete latent variables for other tasks. Given the low perplexity achieved with our method despite not including the words such as articles, neural language model can extract necessary information regardless of human-readability of the context.      

Our approach of compressing a sequence based on token-level losses is related to \citep{sch}, in which the compressed sequence consists of the tokens that the model predicted incorrectly.   \citep{miao} applied VAE to compress and decompress texts just as \citep{kaiser1,roy} and our work. The difference is that the training of their model is semi-supervised, which requires the pairs of an original sentence and its summary, whereas other works mentioned here work in a completely unsupervised manner, which is necessary for scalability, since the size of such pairs tend to be smaller than the original dataset size by an order of magnitude. Notably, it takes a large number of pairs for the model of \citep{miao} to require before achieving a respectable degree of perplexity reduction.

\section{Methods}
\subsection{Motivations}
Not all tokens are created equal. When LM is trained on text, some tokens have significantly higher loss than the other tokens. For example, 'consolidation' is generally harder to predict than 'a'. However, this phenomenon is dependent on the context. Depending on the context available, 'apple' can be harder to predict than 'neural', or vice versa. Thus, a natural approach to compress a sentence is to extract the tokens with highest loss from LM. 

One can reach to the aforementioned argument through trying to solve the following problem that is a generalization of what \citep{kaiser1} tried to achieve: Let $f_{\phi}$ be an algorithm (not necessarily a neural network) that compresses a sequence in terms of its length by a factor of $K$. The size of target vocabulary should be roughly comparable to that of the original vocabulary for practical purpose. Consider a Seq2Seq model $p_{\theta}$ and dataset $D$. Define the conditional loss
\[-\mathbb{E}_{x\sim D}\frac{1}{l_x}\sum_{i=1}^{l_x}\log(\tilde{p}_{\theta}(x_i|c_i,f(x))),\] where $x=(x_1,\ldots x_{l_x})$, $c_i=(x_1,\ldots x_{i-1})$ and $x_i$ is a token. The objective is to minimize the above loss with respect to $(\theta,\phi)$ in hope that the conditional test loss would be minimized. In the case of \citep{kaiser1}, $f_{\phi}$ is a neural network trained concurrently with $p_{\theta}$, and there is no restriction on the discrete latent variables $f_{\phi}(x)$. With discrete latent variable models such as VQ-VAE and Improved Semantic Hashing one can apply SGD with respect to $\theta$ and $\phi$ concurrently, which is seemingly the best method for this problem.

One would naively guess that, if $f_{\phi}(x)$ is constrained to be a subsequence of $x$, the test loss would be smaller than otherwise, since then all the Seq2Seq model has to do is to fill the "gaps" between tokens in the subsequence, which is easier than to encode and decode from scratch. Furthermore, it is also natural to think that the subsequence should be the tokens with highest LM loss, so that the process of filling up the gaps would be much easier for the Seq2Seq model. Concretely, if "I like an apple." is a sentence, and if the LM assigns highest per-token loss to 'like' and 'apple', then we use "like apple" as the context of the sentence. 

In fact, we have observed that our compression method works on the above problem better than state-of-the-art discrete latent variable models, and that the tokens with higher per-token loss given by LM tend to be more informative and useful for reconstruction of the sentence.

Note that the degree of uncertainty for LM to predict the next token tends to be higher at the beginning, since the context available to the model is scarcer at the beginning of the sequence than at the end. This imbalance turns out to be a bottleneck for two reasons. The context created by $f_{\phi}$ has much larger "gaps" for the latter part of the sequence, so the Seq2Seq model predicts less precisely at the later tokens. In addition, it is clearly more advantageous for the model to judge whether a current token is informative or not if the model has more contexts. In order to avoid this imbalance of amount of context information, we need to give the model an access to both future and past context by using bidirectional LM \citep{contextual}, or more specifically bidirectional Transformer \citep{jeremy}, so that at each step the model has access to a whole sequence but the current target.   

\subsection{Proposed Models}

\begin{figure*}[ht]
\vskip 0.2in
\begin{center}
\centerline{\includegraphics[width=0.8\linewidth]{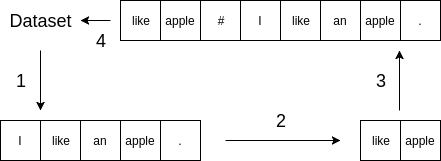}}
\caption{1. Each sample is chosen from the dataset. 2. A score of each token is calculated, and a fixed number of the tokens with the highest scores are selected. For the case of +LM or +bi-LM, the sentence is first evaluated by LM or bi-LM, respectively, and then the loss of each token is evaluated. 3. The selected tokens, along with the separator token, are concatenated to the original sentence with the order unchanged. 4. The original sentence is replaced with the new sentence.}
\label{chart}
\end{center}
\vskip -0.2in
\end{figure*}

Let us denote our model using method X by +X. Since +VQ-VAE is very similar to ISH, we briefly describe its details later. When X is either LM or bi-LM, +X is a LM conditioned on the subsequence of tokens with highest loss as context. For training, we first train the LM or bi-LM and assemble the tokens with highest loss with the order unchanged. Then, we train the LM on the samples with the tokens concatenated as in Figure \ref{chart}.  

We have also attempted +tf-idf and +RL (REINFORCE) as follows. For +tf-idf, instead of the tokens with highest loss the context is a subsequence consisting of the tokens with highest tf-idf scores \citep{tfidf}. For +RL, we let the REINFORCE controller from ENAS \citep{enas} to choose the subsequence of the trained sequence dynamically. The training is essentially identical to that of ENAS, and the controller is trained to maximize the validation perplexity conditioned on the chosen subsequence. The main model (LM) is trained for several epochs, and then the controller is trained. This alternating training continues until the validation log-ppl is maximized. 

To be more clear, +tf-idf, +LM and +bi-LM can be understood as the following. As shown in Figure. \ref{chart}, the original dataset is modified. Then, a vanilla LM is trained and evaluated with this modified dataset just as in the ordinary LM setting.  

+tf-idf is simpler and costs half as much computation as the one with +bi-LM. However, it has at least three short-comings compared with +bi-LM. First, +tf-idf cannot be effectively applied to sequences such as music, since there are many identical tokens, which are treated equally by tf-idf. Secondly, when tf-idf evaluates the importance of a certain token, it pays no attention to the neighboring tokens. The semantic content formed by an ordered set of particular tokens is completely ignored by tf-idf. Lastly, +bi-LM may be more scalable with a larger dataset on which the bi-LM is trained. 

Though our argument has focused on token-level so far, the same can be argued about other hierarchies such as sentence and paragraph. For example, +tf-idf is essentially the reverse version of \citep{wiki} in paragraph-level. The sentence- or token-level analog of our models performs extractive summarization from a story to make its summary and thus makes the text generation of \citep{hier} unsupervised as in \citep{outline}. The compressed sequence can be also considered as the long term policy of sequence generation. Our model learns the long term policy in an unsupervised way. 

\subsection{LEAD Baseline}
\label{lead}
\begin{figure}[ht]
\vskip 0.2in
\begin{center}
\centerline{\includegraphics[width=0.6\columnwidth]{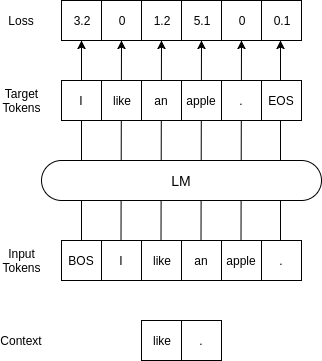}}
\caption{Filtering the loss of the tokens that correspond to each token in the context.}
\label{zero}
\end{center}
\vskip -0.2in
\end{figure}

The trivial baseline we devised is identical to LEAD baseline used in text summarization literature. Let us call this baseline LEAD. The context is a prefix of the sample. With the above example, the model would be conditioned on "I like". Given that the context is already provided, we do not even have to let the model to predict the prefix tokens, as copying the prefix mechanically is a completely valid method for a model. This way, we can treat the loss of the corresponding tokens to be exactly zero. One nice aspect of this simple model is that one can estimate the upper bound of its perplexity from the perplexity of vanilla LM only. The model memorizes the context, so the prediction loss of the prefix that corresponds to the provided context is nearly zero. If the perplexity of first $N$ tokens with LM is also known, one can precisely calculate the perplexity of the LEAD model using the first $N$ tokens as context. The upper bound of the log-ppl of the LEAD baseline is equal to the expectation of the log-ppl by "conditioning on randomly sampled tokens". To make the phrase inside the parenthesis more precise, consider that we randomly sample $N$ distinct integers from $1$ to an integer smaller than the average length of a sequence, say $27$. Now, let us denote the sequence consisting of these sampled integers by $(m_1,\ldots m_N)$ in the ascending order. For each sequence in the dataset, denoted by $T$, we select the first $m_i$-th token $t_i$ for $1\leq i\leq N$. Thus, we have a sequence $(t_1,\ldots t_N)$ for each sequence $T$. Now, if the LM prediction on each sequence $T$ is conditioned on the corresponding subsequence $(t_1,\ldots t_N)$, the log-ppl can be approximately calculated by regarding the perplexity of the conditioned tokens to be zero as in Figure \ref{zero}. 

\subsection{Metric to calculate the compression rate}

\citep{kaiser1} used DSAE (discrete sequence autoencoding efficiency), defined below, to measure the rate of compression achieved by their autoencoder:

\[DSAE :=\frac{K(\log(p) - \log(p'))}{\log(V)}, \]

where $K$, $V$, $\log(p)$ and $\log(p')$ are the length of the original sentence over the length of the compressed sentence, the number of vocabulary of the compressed language, the log-ppl of LM and the log-ppl of the conditional LM, respectively. In fact, DSAE can be larger than $1$, which we believe the authors of \citep{kaiser1} did not intend. This is mostly because, on average, almost half of a minibatch is occupied by zero padding in the setting of 1BLM. \citep{kaiser1} has adjusted the perplexity by ignoring the zero padding tokens, which is a good, common practice. However, for the length of the original sentence, they used the length of the zero-padded sentence instead of the average length of an unpadded sentence, just as in our case, which almost doubled the effective $K$. Therefore, by fixing the value of $K$ and $V$ and using the same dataset, we are going to compare the perplexity to avoid using DSAE except for when comparing our baseline with that of \citep{kaiser1}. 
 
\subsection{Calculation of log-ppl for the LEAD baseline}
\label{calc}
\citep{kaiser1} used a subword tokenizer to fix the total number of vocabulary 32,000 for 1BLM, which we assume extended the average length of a sentence (27) and the length of padded sentence (50) by a certain factor $k$. So, they are now $27k$ and $50k$, respectively. Note that the per-token log-ppl decreases from the beginning of sentence to the end, and this trend is remarkable especially at the beginning. Hence, we can assume the per-token log-ppl at the beginning is much higher than the average per-token log-ppl, which is $3.59$ \citep{kaiser1}. 

On the other hand, since $K=\lceil \frac{50k}{7k}\rceil=8$, the number of the tokens that can be used in the context is approximately $7k$. So to calculate the log-ppl of LEAD baseline, we can instead consider the vanilla LM by setting its per-token log-ppl for the first $7k$ tokens to be zero as in Figure \ref{zero}. Since the per-token log-ppl at the beginning is much higher than $3.59$, the log-ppl of LEAD baseline is calculated to be lower than $\frac{(27k-7k)3.59}{27k}=2.66$. Therefore, our DSAE is greater than $\frac{8(3.59-2.66)}{\log(32000)}=0.72$. In Table \ref{1blm}, we compare the test log-ppl and DSAE on 1BLM among LM, our LEAD baseline and Improved Semantic Hashing. Note that we fix $K=8$. As the table suggests, ISH cannot surpass this baseline. 

In fact, one can easily verify that the expectation of the log-ppl conditioned on the randomly sampled tokens is roughly equal to this upper bound of log-ppl of LEAD. Hence, the fact that ISH cannot even outperform the upper bound of log-ppl of LEAD implies that it cannot even outperform the conditioning on randomly sampled tokens in the sense of Section \ref{lead}.   

\begin{table}[t]
\caption{Test log-ppl on 1BLM.}
\label{1blm}
\vskip 0.15in
\begin{center}
\begin{small}
\begin{sc}
\begin{tabular}{lccr}
Model & log-ppl & DSAE\\
LM &  3.59  & 0 \\
LEAD & $<$2.66 & $>$0.72 \\
ISH \citep{kaiser1} & 2.82  & 0.55\\
\end{tabular}
\end{sc}
\end{small}
\end{center}
\vskip -0.1in
\end{table}

\subsection{Our bi-LM architecture} \label{arch}
\begin{figure*}[ht]
\vskip 0.2in
\begin{center}
\centerline{\includegraphics[width=0.7\linewidth]{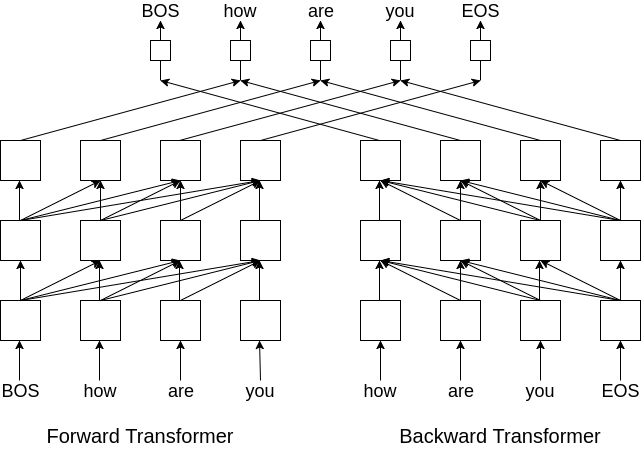}}
\caption{Bidirectional LM Transformer. Note that the number of layers used in our experiments is 6 instead of 3.}
\label{bi}
\end{center}
\vskip -0.2in
\end{figure*}

We have used an architecture similar to ELMo, but we have not used the full power of ELMo. As the Figure \ref{bi} shows, each Transformer processes input completely independently to each other until before the softmax layer. Before the softmax layer is applied, the output from each Transformer is concatenated over the hidden dimension. Then, the softmax layer is applied to predict the target symbol. Each Transformer does not know either past or future context, so the prediction of the next (either immediate future or past) token becomes a non-trivial task. All the other additional components of ELMo was not used in our paper.

\section{Experiments}

In this section, unless specified otherwise, we do not use any regularization method, which often did not result in any improvement in our case. The architecture and hyperparameters of Transformer is identical to that of base Transformer in \citep{tra} except that the filter size is $1024$, not $2048$. Each minibatch consists of 256 sentences from the dataset.

To demonstrate improvement in perplexity, we use EMNLP2017 WMT Dataset, which was preprocessed in \citep{leakgan} to eliminate abnormal sentences. The dataset and the code for preprocessing are publicly available \footnote{\url{https://github.com/geek-ai/Texygen}}. After preprocessed, the dataset contains 5,742 words and about 280,000 sentences for training and 10,000 for testing. The average number of words per sample is 27, and the maximum words per sample is 51, whereas in 1BLM they are 27 and (curtailed to) 50, respectively. For this reason as well as the fact that both of them are a news dataset, WMT can be considered as a smaller variant of 1BLM. For all the experiments using WMTNews, we fix $K=8$, i.e., the number of tokens in the context is $\lceil\frac{51}{8}\rceil=7$. Also, $V$ for WMTNews experiments is similar for both +VQ-VAE ($V=2^{12}$) and other cases ($V=5,742$). Hence, it suffices to compare their perplexity for evaluation of compression rate. 

According to the public repositories of GAN-based text generation models \citep{texy,cot,leakgan}, they used a slightly different definition of log-ppl that treats the zero-padding tokens as ordinary tokens. Let us call this unadjusted log-ppl. Since it is almost trivial to predict the zero-padding token, this results in underestimation of the actual log-ppl. We use an adjusted log-ppl, or per-token log-ppl that ignores the zero-padding token, which is commonly used in other areas. Unless specified otherwise, "log-ppl" refers to the latter. For the case of WMT, the adjusted log-ppl is equal to unadjusted log-ppl multipled by $1.88$.

First, we compare the log-ppl of LSTM, CoT and Transformer as shown in Table \ref{nodlv}. The result demonstrates the scalability of LM, or more specifically MLE training, and the power of Transformer. The log-ppl for LSTM and Transformer was obtained under MLE training. The log-ppl for CoT is cited from \citep{cot}. Note that, for CoT, neither increasing the hidden dimension nor replacing the architecture with Transformer, while trying various other hyperparameters, resulted in a better log-ppl. The quantity inside the parenthesis was obtained when each minibatch contains four times more sentences.   

Then, we compare the log-ppl of LM equipped with VQ-VAE and our models as shown in Table \ref{dlv}. Baseline refers to the Transformer LM in Table \ref{nodlv}. For +VQ-VAE, we used the architecture of Improved Semantic Hashing \citep{kaiser1}, except that the discretization component of Improved Semantic Hashing is replaced with that of VQ-VAE, and that the hyperparameters are searched to achieve the best performance (we obtained a DSAE comparable  to that of \citep{kaiser1}). In particular, we used hard EM instead of soft EM, as the latter requires 10 times as many discrete latent variables as the former, which means soft EM is not suitable for compression. The LM used in +LM is also identical to Transformer in Baseline. The bi-LM used in +bi-LM consists of two Transformers, each of which is identical to the Transformer in Baseline, and was explained in details in Section \ref{arch}. The controller used in +RL is identical to that of \citep{enas}, and so are its hyperparameters. We calculated the upper bound of log-ppl of LEAD on WMTNews just as in Section \ref{calc}: $\frac{(27-7)3.84}{27}=2.84$. To make sure that the bound is correct, we have trained LEAD, which resulted in the perplexity of 2.71 and agree with our calculation.

\begin{table}[t]
\caption{Adjusted and unadjusted test log-ppl.}
\label{nodlv}
\vskip 0.15in
\begin{center}
\begin{small}
\begin{sc}
\begin{tabular}{lcccr}
Model & No adj. & Adj. \\
LSTM. & 2.39  & 4.49 \\ 
CoT  & 2.14 \citep{cot} &  4.02 \\ 
Transformer & 2.04 (1.91) & 3.84 (3.59)\\
\end{tabular}
\end{sc}
\end{small}
\end{center}
\vskip -0.1in
\end{table}

\begin{table}[t]
\caption{Adjusted test log-ppl.}
\label{dlv}
\vskip 0.15in
\begin{center}
\begin{small}
\begin{sc}
\begin{tabular}{lccr}
Model & log-ppl\\
Baseline & 3.84\\
+VQ-VAE & 3.08 \\
LEAD & 2.71 ($<2.84$) \\
+LM & 2.61 \\
+bi-LM & 2.32 \\
+tf-idf & 2.36 \\
+RL & 3.02 \\
\end{tabular}
\end{sc}
\end{small}
\end{center}
\vskip -0.1in
\end{table}

As Table \ref{dlv} shows, +tf-idf and +bi-LM perform the best, far better than +VQ-VAE. Notably, +RL performs only as well as +VQ-VAE. We observed that the distribution of average per-token loss over length dimension for bi-LM is almost uniform except at the both edges. Furthermore, the subsequence generated by +tf-idf is similar to that of +bi-LM, which is unsurprising given their small gap in performance. 

\section{Discussion and Future Directions}
It is not generally true that conditional text generation model with lower ppl results in better generation quality. Hence, it is not a priori clear whether achieving lower conditional perplexity by optimizing $f_{\phi}$ and restricting $f_{\phi}(x)$ to be a subsequence of $x$ improves the quality of the sequence generated by hierarchical generation. For example, if the context consists of tokens with high log-ppl, then generating the context with LM must be harder; therefore, one can argue that hierarchical generation with our models may not improve the generation quality. We have some counterargument to this claim. Based on our arguments and experiments, there is an irrefutable similarity between extractive summarization and how optimized $f_{\phi}$ behaves. This means that hierarchical generation with our method is nothing but an unsupervised analog of \citep{hier} hence should should perform likewise. 

From our result, we hypothesize that LM prediction implicitly performs a sophisticated information retrieval that is sensitive to informative keywords. Given that +bi-LM that achieved the log-ppl of 2.32 outperforms over randomly selecting tokens that achieved the log-ppl of 2.84, by the difference in log-ppl as large as 0.52, +bi-LM (and +tf-idf) is very sensitive to informative tokens, and the most informative tokens contain disproportionately greater information than the rest. In other words, there is a small number of words that contains a great deal of information of the remaining part of the text. In fact, conditioning on the extracted tokens does not only reduce the loss of the tokens in the context but also other tokens. Since our methods are purely based on ranking with individual token's score, it is reasonable to assume that there is a following chain of subsequences: $()=C_0 < C_1\ldots <C_T$, where $C_n$ is a subsequence with length $n$, $C_T$ is the original text, and the LM of $C_T$ conditioned on $C_n$ gives a log-ppl close to the best possible log-ppl conditional on any subsequence of $C_T$ with the same length. This kind of hierarchy of informativeness in text is what we believe to be the key for language to be a great compression code.  

For the near future, we should seek for extractive, not abstractive, methods for compression. While it is possible that in the future there will be an unsupervised abstractive method that fulfills our needs, it may be easier to lower the perplexity significantly by simply refining extractive methods. In fact, \citep{wiki} showed that with their state-of-the-art abstractive summarization algorithm, which consists of an extractive and a subsequent abstractive process, they obtained a reduction in perplexity by $53\%$ when the extractive method (tf-idf) is replaced with an oracle method that has an access to the target information. Therefore, we believe that there is a significant potential for unsupervised extractive methods in improving the compression rate further. 

We believe that language is also a strong compression code for data of modality other than text. For example, a small bytes of words can semantically pertinently express objects that can barely be expressed by millions of pixels. Let us consider the case where we apply our method to the setting of Image Transformer \citep{image} for simplicity. Since the sum of per-pixel NLL over all the pixels on a single image is far greater than the sum of per-token NLL over all the tokens of their summary sentence, the perplexity reduction by conditional generation is minimal. However, as reported by \citep{image}, generation conditional on mere class label visibly improved the generation quality. If we regard class label as a form of summary, this suggests that image analog of our method should work well. However, in this case perplexity should be replaced with some metric that better captures the degree of improvement in image quality. Also, some data such as music cannot be naturally summarized with text. However, our models may be able to summarize such data with good compression rate. We try to investigate these problems further in our future study. 

%\subsubsection*{Acknowledgments}

\bibliography{iclr2019_conference}

\begin{thebibliography}{40}
\providecommand{\natexlab}[1]{#1}
\providecommand{\url}[1]{\texttt{#1}}
\expandafter\ifx\csname urlstyle\endcsname\relax
  \providecommand{\doi}[1]{doi: #1}\else
  \providecommand{\doi}{doi: \begingroup \urlstyle{rm}\Url}\fi

\bibitem[{Baevski} \& {Auli}(2018){Baevski} and {Auli}]{adap}
A.~{Baevski} and M.~{Auli}.
\newblock {Adaptive Input Representations for Neural Language Modeling}.
\newblock \emph{ArXiv e-prints}, September 2018.

\bibitem[{Bowman} et~al.(2015){Bowman}, {Vilnis}, {Vinyals}, {Dai},
  {Jozefowicz}, and {Bengio}]{cont}
S.~R. {Bowman}, L.~{Vilnis}, O.~{Vinyals}, A.~M. {Dai}, R.~{Jozefowicz}, and
  S.~{Bengio}.
\newblock {Generating Sentences from a Continuous Space}.
\newblock \emph{ArXiv e-prints}, November 2015.

\bibitem[{Caccia} et~al.(2018){Caccia}, {Caccia}, {Fedus}, {Larochelle},
  {Pineau}, and {Charlin}]{falling}
Massimo {Caccia}, Lucas {Caccia}, William {Fedus}, Hugo {Larochelle}, Joelle
  {Pineau}, and Laurent {Charlin}.
\newblock {Language GANs Falling Short}.
\newblock \emph{arXiv e-prints}, art. arXiv:1811.02549, November 2018.

\bibitem[{Chelba} et~al.(2013){Chelba}, {Mikolov}, {Schuster}, {Ge}, {Brants},
  {Koehn}, and {Robinson}]{1b}
C.~{Chelba}, T.~{Mikolov}, M.~{Schuster}, Q.~{Ge}, T.~{Brants}, P.~{Koehn}, and
  T.~{Robinson}.
\newblock {One Billion Word Benchmark for Measuring Progress in Statistical
  Language Modeling}.
\newblock \emph{ArXiv e-prints}, December 2013.

\bibitem[{Dai} et~al.(2019){Dai}, {Yang}, {Yang}, {Carbonell}, {Le}, and
  {Salakhutdinov}]{xl}
Zihang {Dai}, Zhilin {Yang}, Yiming {Yang}, Jaime {Carbonell}, Quoc~V. {Le},
  and Ruslan {Salakhutdinov}.
\newblock {Transformer-XL: Attentive Language Models Beyond a Fixed-Length
  Context}.
\newblock \emph{arXiv e-prints}, art. arXiv:1901.02860, January 2019.

\bibitem[{Daniluk} et~al.(2017){Daniluk}, {Rockt{\"a}schel}, {Welbl}, and
  {Riedel}]{frust}
Micha{\l} {Daniluk}, Tim {Rockt{\"a}schel}, Johannes {Welbl}, and Sebastian
  {Riedel}.
\newblock {Frustratingly Short Attention Spans in Neural Language Modeling}.
\newblock \emph{arXiv e-prints}, art. arXiv:1702.04521, February 2017.

\bibitem[{Devlin} et~al.(2018){Devlin}, {Chang}, {Lee}, and {Toutanova}]{mask}
J.~{Devlin}, M.-W. {Chang}, K.~{Lee}, and K.~{Toutanova}.
\newblock {BERT: Pre-training of Deep Bidirectional Transformers for Language
  Understanding}.
\newblock \emph{ArXiv e-prints}, October 2018.

\bibitem[{Dieleman} et~al.(2018){Dieleman}, {van den Oord}, and
  {Simonyan}]{music}
S.~{Dieleman}, A.~{van den Oord}, and K.~{Simonyan}.
\newblock {The challenge of realistic music generation: modelling raw audio at
  scale}.
\newblock \emph{ArXiv e-prints}, June 2018.

\bibitem[{Drissi} et~al.(2018){Drissi}, {Watkins}, and {Kalita}]{outline}
M.~{Drissi}, O.~{Watkins}, and J.~{Kalita}.
\newblock {Hierarchical Text Generation using an Outline}.
\newblock \emph{ArXiv e-prints}, October 2018.

\bibitem[Fan et~al.(2018)Fan, Lewis, and Dauphin]{hier}
Angela Fan, Mike Lewis, and Yann Dauphin.
\newblock Hierarchical neural story generation.
\newblock \emph{CoRR}, abs/1805.04833, 2018.
\newblock URL \url{http://arxiv.org/abs/1805.04833}.

\bibitem[{Fedus} et~al.(2018){Fedus}, {Goodfellow}, and {Dai}]{maskgan}
W.~{Fedus}, I.~{Goodfellow}, and A.~M. {Dai}.
\newblock {MaskGAN: Better Text Generation via Filling in the\_\_\_\_\_\_}.
\newblock \emph{ArXiv e-prints}, January 2018.

\bibitem[{Gulrajani} et~al.(2017){Gulrajani}, {Ahmed}, {Arjovsky}, {Dumoulin},
  and {Courville}]{wgan-gp}
I.~{Gulrajani}, F.~{Ahmed}, M.~{Arjovsky}, V.~{Dumoulin}, and A.~{Courville}.
\newblock {Improved Training of Wasserstein GANs}.
\newblock \emph{ArXiv e-prints}, March 2017.

\bibitem[Guo et~al.(2017)Guo, Lu, Cai, Zhang, Yu, and Wang]{leakgan}
Jiaxian Guo, Sidi Lu, Han Cai, Weinan Zhang, Yong Yu, and Jun Wang.
\newblock Long text generation via adversarial training with leaked
  information.
\newblock \emph{CoRR}, abs/1709.08624, 2017.
\newblock URL \url{http://arxiv.org/abs/1709.08624}.

\bibitem[{Hassan} et~al.(2018){Hassan}, {Aue}, {Chen}, {Chowdhary}, {Clark},
  {Federmann}, {Huang}, {Junczys-Dowmunt}, {Lewis}, {Li}, {Liu}, {Liu}, {Luo},
  {Menezes}, {Qin}, {Seide}, {Tan}, {Tian}, {Wu}, {Wu}, {Xia}, {Zhang},
  {Zhang}, and {Zhou}]{parity}
H.~{Hassan}, A.~{Aue}, C.~{Chen}, V.~{Chowdhary}, J.~{Clark}, C.~{Federmann},
  X.~{Huang}, M.~{Junczys-Dowmunt}, W.~{Lewis}, M.~{Li}, S.~{Liu}, T.-Y. {Liu},
  R.~{Luo}, A.~{Menezes}, T.~{Qin}, F.~{Seide}, X.~{Tan}, F.~{Tian}, L.~{Wu},
  S.~{Wu}, Y.~{Xia}, D.~{Zhang}, Z.~{Zhang}, and M.~{Zhou}.
\newblock {Achieving Human Parity on Automatic Chinese to English News
  Translation}.
\newblock \emph{ArXiv e-prints}, March 2018.

\bibitem[{Hestness} et~al.(2017){Hestness}, {Narang}, {Ardalani}, {Diamos},
  {Jun}, {Kianinejad}, {Patwary}, {Yang}, and {Zhou}]{scaling}
J.~{Hestness}, S.~{Narang}, N.~{Ardalani}, G.~{Diamos}, H.~{Jun},
  H.~{Kianinejad}, M.~M.~A. {Patwary}, Y.~{Yang}, and Y.~{Zhou}.
\newblock {Deep Learning Scaling is Predictable, Empirically}.
\newblock \emph{ArXiv e-prints}, December 2017.

\bibitem[{Howard} \& {Ruder}(2018){Howard} and {Ruder}]{jeremy}
J.~{Howard} and S.~{Ruder}.
\newblock {Universal Language Model Fine-tuning for Text Classification}.
\newblock \emph{ArXiv e-prints}, January 2018.

\bibitem[{Hu} et~al.(2017){Hu}, {Yang}, {Liang}, {Salakhutdinov}, and
  {Xing}]{control}
Z.~{Hu}, Z.~{Yang}, X.~{Liang}, R.~{Salakhutdinov}, and E.~P. {Xing}.
\newblock {Toward Controlled Generation of Text}.
\newblock \emph{ArXiv e-prints}, March 2017.

\bibitem[Kaiser \& Bengio(2018)Kaiser and Bengio]{kaiser1}
Lukasz Kaiser and Samy Bengio.
\newblock Discrete autoencoders for sequence models.
\newblock \emph{CoRR}, abs/1801.09797, 2018.
\newblock URL \url{http://arxiv.org/abs/1801.09797}.

\bibitem[Kaiser et~al.(2018)Kaiser, Roy, Vaswani, Parmar, Bengio, Uszkoreit,
  and Shazeer]{kaiser2}
Lukasz Kaiser, Aurko Roy, Ashish Vaswani, Niki Parmar, Samy Bengio, Jakob
  Uszkoreit, and Noam Shazeer.
\newblock Fast decoding in sequence models using discrete latent variables.
\newblock \emph{CoRR}, abs/1803.03382, 2018.
\newblock URL \url{http://arxiv.org/abs/1803.03382}.

\bibitem[{Liu} et~al.(2018){Liu}, {Saleh}, {Pot}, {Goodrich}, {Sepassi},
  {Kaiser}, and {Shazeer}]{wiki}
P.~J. {Liu}, M.~{Saleh}, E.~{Pot}, B.~{Goodrich}, R.~{Sepassi}, L.~{Kaiser},
  and N.~{Shazeer}.
\newblock {Generating Wikipedia by Summarizing Long Sequences}.
\newblock \emph{ArXiv e-prints}, January 2018.

\bibitem[{Lu} et~al.(2018){Lu}, {Yu}, {Zhang}, and {Yu}]{cot}
S.~{Lu}, L.~{Yu}, W.~{Zhang}, and Y.~{Yu}.
\newblock {CoT: Cooperative Training for Generative Modeling}.
\newblock \emph{ArXiv e-prints}, April 2018.

\bibitem[{Menick} \& {Kalchbrenner}(2018){Menick} and {Kalchbrenner}]{spn}
Jacob {Menick} and Nal {Kalchbrenner}.
\newblock {Generating High Fidelity Images with Subscale Pixel Networks and
  Multidimensional Upscaling}.
\newblock \emph{arXiv e-prints}, art. arXiv:1812.01608, December 2018.

\bibitem[{Miao} \& {Blunsom}(2016){Miao} and {Blunsom}]{miao}
Yishu {Miao} and Phil {Blunsom}.
\newblock {Language as a Latent Variable: Discrete Generative Models for
  Sentence Compression}.
\newblock \emph{arXiv e-prints}, art. arXiv:1609.07317, September 2016.

\bibitem[{Ott} et~al.(2018){Ott}, {Auli}, {Grangier}, and
  {Ranzato}]{uncertainty}
M.~{Ott}, M.~{Auli}, D.~{Grangier}, and M.~{Ranzato}.
\newblock {Analyzing Uncertainty in Neural Machine Translation}.
\newblock \emph{ArXiv e-prints}, February 2018.

\bibitem[{Parmar} et~al.(2018){Parmar}, {Vaswani}, {Uszkoreit}, {Kaiser},
  {Shazeer}, {Ku}, and {Tran}]{image}
Niki {Parmar}, Ashish {Vaswani}, Jakob {Uszkoreit}, {\L}ukasz {Kaiser}, Noam
  {Shazeer}, Alexander {Ku}, and Dustin {Tran}.
\newblock {Image Transformer}.
\newblock \emph{arXiv e-prints}, art. arXiv:1802.05751, February 2018.

\bibitem[{Peters} et~al.(2018){Peters}, {Neumann}, {Iyyer}, {Gardner}, {Clark},
  {Lee}, and {Zettlemoyer}]{contextual}
M.~E. {Peters}, M.~{Neumann}, M.~{Iyyer}, M.~{Gardner}, C.~{Clark}, K.~{Lee},
  and L.~{Zettlemoyer}.
\newblock {Deep contextualized word representations}.
\newblock \emph{ArXiv e-prints}, February 2018.

\bibitem[{Pham} et~al.(2018){Pham}, {Guan}, {Zoph}, {Le}, and {Dean}]{enas}
H.~{Pham}, M.~Y. {Guan}, B.~{Zoph}, Q.~V. {Le}, and J.~{Dean}.
\newblock {Efficient Neural Architecture Search via Parameter Sharing}.
\newblock \emph{ArXiv e-prints}, February 2018.

\bibitem[{Press} et~al.(2017){Press}, {Bar}, {Bogin}, {Berant}, and
  {Wolf}]{rnn}
O.~{Press}, A.~{Bar}, B.~{Bogin}, J.~{Berant}, and L.~{Wolf}.
\newblock {Language Generation with Recurrent Generative Adversarial Networks
  without Pre-training}.
\newblock \emph{ArXiv e-prints}, June 2017.

\bibitem[Ramos(2003)]{tfidf}
Juan~Enrique Ramos.
\newblock Using tf-idf to determine word relevance in document queries.
\newblock 2003.

\bibitem[Roy et~al.(2018)Roy, Vaswani, Neelakantan, and Parmar]{roy}
Aurko Roy, Ashish Vaswani, Arvind Neelakantan, and Niki Parmar.
\newblock Theory and experiments on vector quantized autoencoders.
\newblock \emph{CoRR}, abs/1805.11063, 2018.
\newblock URL \url{http://arxiv.org/abs/1805.11063}.

\bibitem[Schmidhuber(1992)]{sch}
J\"{u}rgen Schmidhuber.
\newblock Learning complex, extended sequences using the principle of history
  compression.
\newblock \emph{Neural Comput.}, 4\penalty0 (2):\penalty0 234--242, March 1992.
\newblock ISSN 0899-7667.
\newblock \doi{10.1162/neco.1992.4.2.234}.
\newblock URL \url{http://dx.doi.org/10.1162/neco.1992.4.2.234}.

\bibitem[{Semeniuta} et~al.(2018){Semeniuta}, {Severyn}, and {Gelly}]{eval}
S.~{Semeniuta}, A.~{Severyn}, and S.~{Gelly}.
\newblock {On Accurate Evaluation of GANs for Language Generation}.
\newblock \emph{ArXiv e-prints}, June 2018.

\bibitem[{Silver} et~al.(2017){Silver}, {Hubert}, {Schrittwieser},
  {Antonoglou}, {Lai}, {Guez}, {Lanctot}, {Sifre}, {Kumaran}, {Graepel},
  {Lillicrap}, {Simonyan}, and {Hassabis}]{alphazero}
D.~{Silver}, T.~{Hubert}, J.~{Schrittwieser}, I.~{Antonoglou}, M.~{Lai},
  A.~{Guez}, M.~{Lanctot}, L.~{Sifre}, D.~{Kumaran}, T.~{Graepel},
  T.~{Lillicrap}, K.~{Simonyan}, and D.~{Hassabis}.
\newblock {Mastering Chess and Shogi by Self-Play with a General Reinforcement
  Learning Algorithm}.
\newblock \emph{ArXiv e-prints}, December 2017.

\bibitem[van~den Oord et~al.(2017)van~den Oord, Vinyals, and
  Kavukcuoglu]{vqvae}
A{\"{a}}ron van~den Oord, Oriol Vinyals, and Koray Kavukcuoglu.
\newblock Neural discrete representation learning.
\newblock \emph{CoRR}, abs/1711.00937, 2017.
\newblock URL \url{http://arxiv.org/abs/1711.00937}.

\bibitem[{Vaswani} et~al.(2017){Vaswani}, {Shazeer}, {Parmar}, {Uszkoreit},
  {Jones}, {Gomez}, {Kaiser}, and {Polosukhin}]{tra}
A.~{Vaswani}, N.~{Shazeer}, N.~{Parmar}, J.~{Uszkoreit}, L.~{Jones}, A.~N.
  {Gomez}, L.~{Kaiser}, and I.~{Polosukhin}.
\newblock {Attention Is All You Need}.
\newblock \emph{ArXiv e-prints}, June 2017.

\bibitem[{Wu} et~al.(2016){Wu}, {Schuster}, {Chen}, {Le}, {Norouzi},
  {Macherey}, {Krikun}, {Cao}, {Gao}, {Macherey}, {Klingner}, {Shah},
  {Johnson}, {Liu}, {Kaiser}, {Gouws}, {Kato}, {Kudo}, {Kazawa}, {Stevens},
  {Kurian}, {Patil}, {Wang}, {Young}, {Smith}, {Riesa}, {Rudnick}, {Vinyals},
  {Corrado}, {Hughes}, and {Dean}]{google}
Y.~{Wu}, M.~{Schuster}, Z.~{Chen}, Q.~V. {Le}, M.~{Norouzi}, W.~{Macherey},
  M.~{Krikun}, Y.~{Cao}, Q.~{Gao}, K.~{Macherey}, J.~{Klingner}, A.~{Shah},
  M.~{Johnson}, X.~{Liu}, {\L}.~{Kaiser}, S.~{Gouws}, Y.~{Kato}, T.~{Kudo},
  H.~{Kazawa}, K.~{Stevens}, G.~{Kurian}, N.~{Patil}, W.~{Wang}, C.~{Young},
  J.~{Smith}, J.~{Riesa}, A.~{Rudnick}, O.~{Vinyals}, G.~{Corrado},
  M.~{Hughes}, and J.~{Dean}.
\newblock {Google's Neural Machine Translation System: Bridging the Gap between
  Human and Machine Translation}.
\newblock \emph{ArXiv e-prints}, September 2016.

\bibitem[Yu et~al.(2016)Yu, Zhang, Wang, and Yu]{seqgan}
Lantao Yu, Weinan Zhang, Jun Wang, and Yong Yu.
\newblock Seqgan: Sequence generative adversarial nets with policy gradient.
\newblock \emph{CoRR}, abs/1609.05473, 2016.
\newblock URL \url{http://arxiv.org/abs/1609.05473}.

\bibitem[{Zhang} et~al.(2017){Zhang}, {Gan}, {Fan}, {Chen}, {Henao}, {Shen},
  and {Carin}]{feature}
Y.~{Zhang}, Z.~{Gan}, K.~{Fan}, Z.~{Chen}, R.~{Henao}, D.~{Shen}, and
  L.~{Carin}.
\newblock {Adversarial Feature Matching for Text Generation}.
\newblock \emph{ArXiv e-prints}, June 2017.

\bibitem[Zhao et~al.(2018)Zhao, Luo, and Aizawa]{compr}
Yang Zhao, Zhiyuan Luo, and Akiko Aizawa.
\newblock A language model based evaluator for sentence compression.
\newblock In \emph{Proceedings of the 56th Annual Meeting of the Association
  for Computational Linguistics (Volume 2: Short Papers)}, pp.\  170--175.
  Association for Computational Linguistics, 2018.
\newblock URL \url{http://aclweb.org/anthology/P18-2028}.

\bibitem[{Zhu} et~al.(2018){Zhu}, {Lu}, {Zheng}, {Guo}, {Zhang}, {Wang}, and
  {Yu}]{texy}
Y.~{Zhu}, S.~{Lu}, L.~{Zheng}, J.~{Guo}, W.~{Zhang}, J.~{Wang}, and Y.~{Yu}.
\newblock {Texygen: A Benchmarking Platform for Text Generation Models}.
\newblock \emph{ArXiv e-prints}, February 2018.

\end{thebibliography}
\bibliographystyle{iclr2019_conference}

\end{document}